\documentclass[conference]{IEEEtran}
\IEEEoverridecommandlockouts
\usepackage{cite}

\usepackage{amsmath,amssymb,amsfonts}
\usepackage{algorithmic}
\usepackage{graphicx}
\usepackage{textcomp}
\usepackage{xcolor}
\usepackage{float}
\def\BibTeX{{\rm B\kern-.05em{\sc i\kern-.025em b}\kern-.08em
    T\kern-.1667em\lower.7ex\hbox{E}\kern-.125emX}}

\begin{document}

\title{Olive Tree Satellite Image Segmentation Based On SAM and Multi-Phase Refinement\\}

\author{
\IEEEauthorblockN{1\textsuperscript{st} Amir Jmal}
\IEEEauthorblockA{
\textit{National School of Electronics and Telecomms of Sfax}\\
Sfax, Tunisia \\
amir.jmal16@gmail.com}
\and
\IEEEauthorblockN{2\textsuperscript{nd} Chaima Chtourou}
\IEEEauthorblockA{
\textit{National School of Electronics and Telecomms of Sfax}\\
Sfax, Tunisia \\
chaimachtourou1407@gmail.com}
\and
\IEEEauthorblockN{3\textsuperscript{rd} Mahdi Louati}
\IEEEauthorblockA{\textit{SMARTS Laboratory} \\
\textit{National School of Electronics and Telecomms of Sfax}\\
Sfax, Tunisia \\
mahdi.louati@enetcom.usf.tn}
\and
\IEEEauthorblockN{4\textsuperscript{th} Abdelaziz Kallel}
\IEEEauthorblockA{\textit{SMARTS Laboratory} \\
\textit{Digital Research Center of Sfax}\\
Sfax, Tunisia \\
Abdelaziz.Kallel@crns.tn}
\and
\IEEEauthorblockN{5\textsuperscript{th} Houda Khmila}
\IEEEauthorblockA{\textit{Sofrecom Tunisiay} \\
houda.khmila@sofrecom.com}
}

\maketitle

\begin{abstract}
In the context of proven climate change, maintaining olive biodiversity through early anomaly detection and treatment using remote sensing technology is crucial, offering effective management solutions.
This paper presents an innovative approach to olive tree segmentation from satellite images. By leveraging foundational models and advanced segmentation techniques, the study integrates the Segment Anything Model (SAM) to accurately identify and segment olive trees in agricultural plots.
The methodology includes SAM segmentation and corrections based on trees alignement in the field and a learanble constraint about the shape and the size.
Our approach achieved a 98\% accuracy rate, significantly surpassing the initial SAM performance of 82\%.

\end{abstract}


\section{Introduction}

The Tunisian landscape, abundant with olive groves, heavily relies on its agricultural sector, with olive oil being its cornerstone, representing up to 50\% of its exports. In this context, olive tree inventory, growth and health monitoring at large scale appears crucial. For that, olive tree segmentation from satellite images emerges as an important milestone. Indeed, it offers a promising solution for efficiently detecting and monitoring potential anomalies affecting olive trees, thereby contributing to the preservation and optimal management of these agricultural resources.\par

Segmentation is among the main process in image analysis and comprehension, particularly in satellite when it is question about plant detection and monitoring. By dividing images into meaningful segments, it enables more efficient and accurate analysis. This simplification of complex images enhances tasks such as object recognition and tomporal evolution study in case of image time series processing. Consequently, segmentation facilitates precise identification of individual elements within an image, leading to targeted interventions and improved decision-making \cite{segmentation}. \par

In the realm of olive tree segmentation, various approaches have been employed, ranging from traditional machine learning techniques such as K-means clustering \cite{k-means}\cite{waleed2020automatic}, Random Forests \cite{randomForests}, and Support Vector Machines (SVM) \cite{SVM} to more advanced deep learning techniques like U-Net \cite{U-Net}, EfficientDet  \cite{EfficientDet}, SwinTUnet \cite{abozeid2022large}, Detectron2 \cite{abdallah2022olive} and YOLO \cite{YOLO}. These methods have played a significant role in the field, providing valuable insights and solutions. However, the advent of the Segment Anything Model (SAM) \cite{sam}, a foundational model based on generative artificial intelligence, marks a pivotal advancement in this domain. SAM revolutionizes image segmentation by offering a comprehensive and automated solution, transcending the limitations of previous methodologies.\par

The Segment Anything Model (SAM) demonstrates potential in remote sensing applications; however, it has significant limitations, especially when dealing with complex scenarios involving low-resolution images. In such cases, SAM's accuracy decreases, impacting its effectiveness in tasks that require detailed image segmentation. Moreover, while SAM's zero-shot learning capabilities and generalization are beneficial, the model struggles to adapt to specific remote sensing contexts without additional training or fine-tuning \cite{sam-}.\par

Developing a multi-phase approach to effectively segment olive groves is required for overcoming the limitations of SAM. While SAM can perform initial segmentation, its ability to handle complex scenarios, such as those involving low-resolution images or irregular olive formations, remains limited. By integrating a multi-phase approach that takes into account a priori knowledge about the field organization and olive tree shape and form, we can capitalize on SAM's strengths while introducing additional strategies to enhance overall segmentation accuracy. As olive trees often grow in a regular grid pattern, enabling the detection of their positions, we can reduce the complexity of the segmentation problem by focusing on potential regions of interest. By performing segmentation locally on these detected positions, we can further refine the results, taking into account the validity in shapes and sizes of olive trees within a given context.

\section{Theoretical Background}
In our study, we leverage advanced segmentation methods, including the Segment Anything Model (SAM) and its associated classes, such as SAM Automatic Mask Generator (SAMG) and SAM Predictor (SP), along with shape similarity analysis, to address the segmentation of olive trees. 

\subsection{The Segment Anything Model (SAM)}

Developed by Meta AI and launched in April 2023, SAM revolutionizes image segmentation by democratizing this complex task and eliminating the need for manual annotation, making segmentation accessible and efficient. Its open-source nature encourages collaboration and innovation within the AI community. As a foundation model, SAM is pretrained on the SA-1B Dataset, the largest collection of segmentation masks to date, allowing it to identify and segment each object in an image, providing automation and ease of use. SAM’s versatility allows it to detect and segment objects across diverse domains, including underwater, microscopic, aerial, and agricultural contexts \cite{sam}.

\begin{itemize}
\item \textbf{SAM Automatic Mask Generator (SAMG)}
The SAMG class samples pointwise instructions across the image grid to enable SAM to predict multiple masks per point. Masks are then filtered for quality and deduplicated. Additional enhancements are implemented, like post-processing to remove small disjointed regions and holes, further improve mask quality.
\item \textbf{SAM Predictor (SP)}
The SP class efficiently predicts object masks from given instructions by converting images into embeddings. Users can define an image and provide instructions—such as points, boxes, or previous masks—via the predict method to generate high-quality masks. For instance, if a user points to an olive tree in the reference image, the model identifies the tree's features at that point and generates a corresponding mask.
\end{itemize}

\subsection{Shape similarity}
We employed Cosine similarity to evaluate the shape similarity between two segmentation masks. This metric calculates the cosine of the angle between the two vectors, offering a numerical indication of their resemblance. Ranging from -1 (indicating complete dissimilarity) to 1 (representing a perfect match), Cosine similarity finds versatile applications across diverse fields.

The cosine similarity between two vectors \( \mathbf{A} \) et \( \mathbf{B} \) is given by :
\begin{equation}\label{cossim}
\text{CosineSimilarity} = \text{Sc}(\mathbf{A}, \mathbf{B}) = \cos(\theta) = \frac{\mathbf{A} \cdot \mathbf{B}}{\|\mathbf{A}\| \cdot \|\mathbf{B}\|}
\end{equation}

\section{Proposed approach}
\begin{figure*}[t] 
    \centering
    \includegraphics[width=\linewidth]{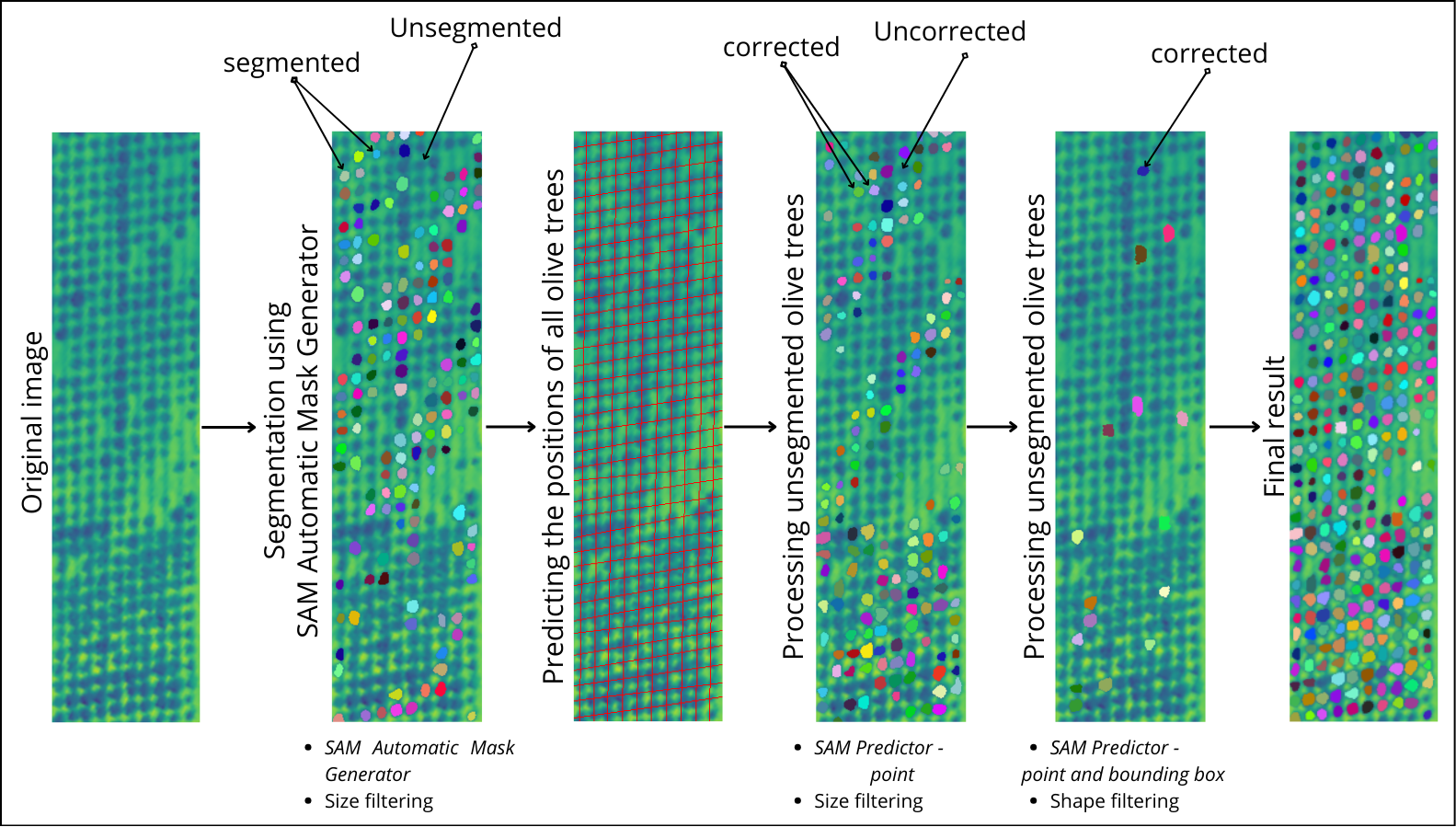} 
    \caption{Overview of the proposed multi-phase approach for olive tree segmentation from satellite images}
    \label{fig:phases}
\end{figure*}

The overview of the multi-phase proposed approach, as illustrated in Fig.~\ref{fig:phases}, begins with segmentation using the SAM Automatic Mask Generator in the first phase. The second phase focuses on predicting the positions of all olive trees by detecting olive tree rows and columns, and thus their intersections. The third phase is dedicated to refining the segmentation of unsegmented olive trees with the SAM Predictor, employing two distinct methods. The first method relies on the centers of unsegmented olive trees, while the second method integrates both the centers and bounding boxes of unsegmented olive trees, aiming for better performance.

\subsection{Phase 1 : Segmentation using SAMG }
In this phase, we use the SAMG class with default settings to process the image and obtain reliable masks. When there are many objects to detect, for instance in a large image, SAM may struggle with precise detection. In such cases, it is necessary to divide the images into smaller sections and apply the generator to each section individually. In our case, when the number of segments is higher than 100, the subdivision processing is applied.

\begin{figure}[H]
    \centering
    \includegraphics[width=1\linewidth]{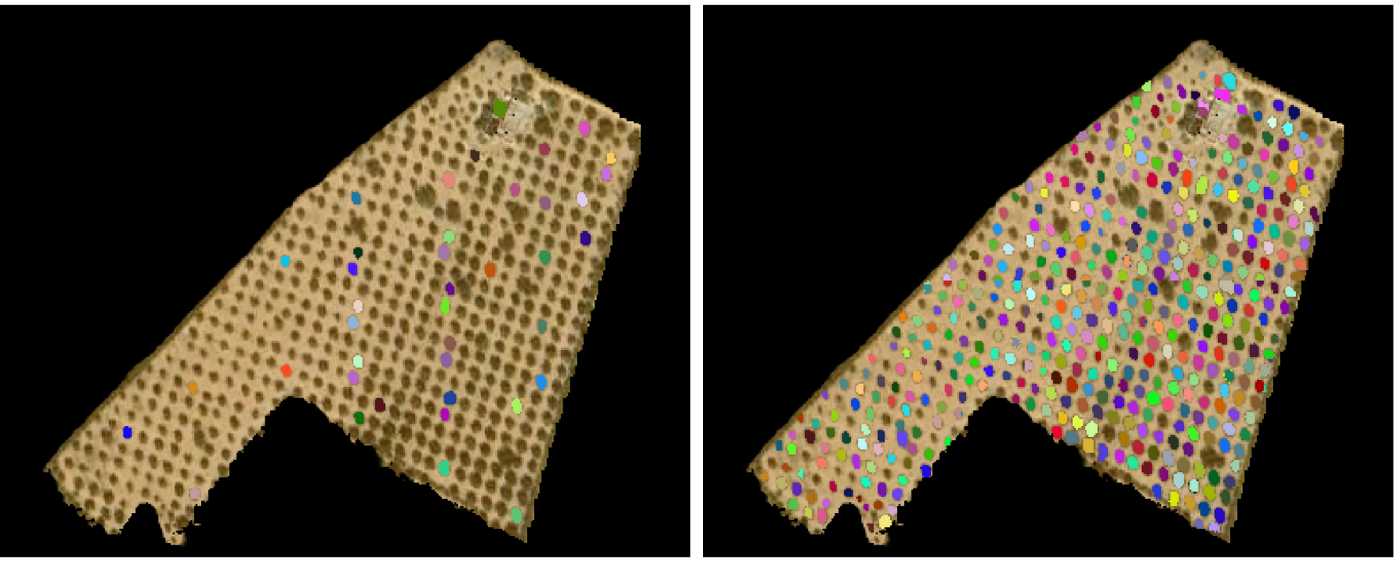}
    \caption{Comparison before and after improvement in olive tree segmentation}
    \label{fig:comparaison}
\end{figure}

In the example shown in Fig.~\ref{fig:comparaison}, the image is 310 x 410 pixels. Subdividing it into smaller parts improved the accuracy of olive tree segmentation.

Once the masks are generated, each is validated using a filtering function. Masks with areas that deviate greatly from the majority are filtered out. The function sets upper and lower limits, determined using the interquartile range (IQR), to identify outliers based on their area.

To determine the positions of segmented olive trees, extract the centers from the validated masks to locate them precisely in the image. In the next phase, these centers will be used to trace rows and columns of olive trees, establishing a reference grid for further analysis.

\subsection{Phase 2 : Predicting the positions of all olive trees}
Rows and columns that define the arrangement of olive trees are created by using the segmented olive tree centers as reference points. The intersections are found to help locate each tree in the image.

\begin{figure}[H]
    \centering
    \includegraphics[width=1\linewidth]{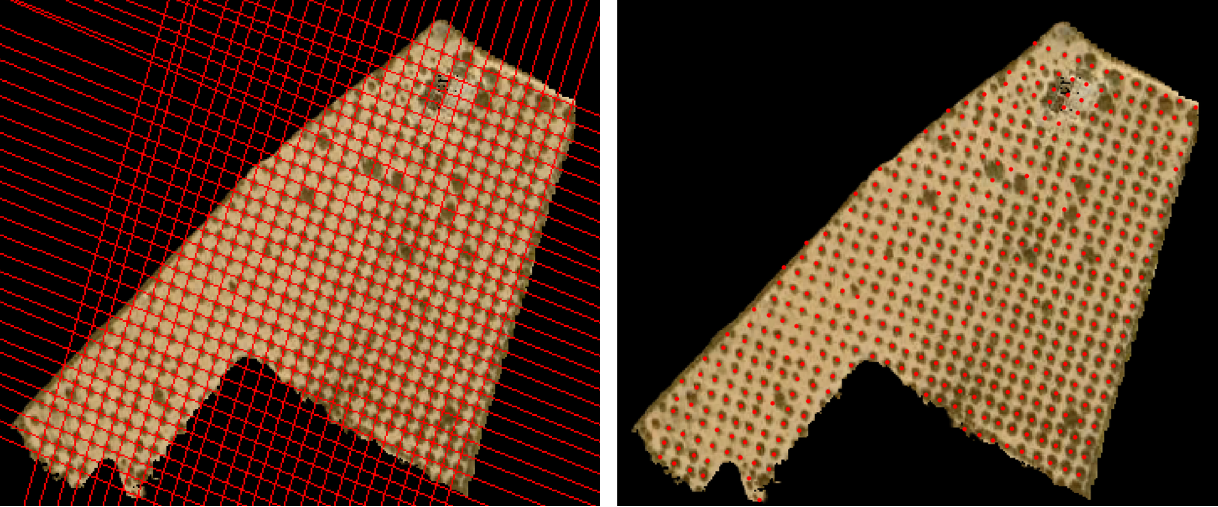}
    \caption{Detection of olive tree rows, columns, and intersections}
    \label{fig:intersections}
\end{figure}

\begin{itemize}
\item \textbf{Detecting olive tree rows}
As olive tree field is organized in a regular grid, therefore if one takes a line parallel to the tree rows then the distance from the trees to this line takes only a few values corresponding to the distance of the different rows to the line. This is not the case if the line is not parallel to the grid and random distance from trees to the line can be found. This fact is used in the following to detect such a parallel line. 

All possible rows are drawn, covering slopes from 0 to 180 degrees to account for all orientations of olive tree rows. Next, the line parallel to the rows is identified by calculating the histogram of distances between each possible row and all segmented olive tree centers. The correct row is chosen based on the maximum variance in the histogram, indicating that points are regularly spaced along it.

To calculate the distance between olive tree rows, we use the correct row's histogram to find peaks that show the positions of olive tree rows. Distances are extracted from each peak to measure the distances between successive peaks and estimate the likely distance.

To trace all correct rows, detected points are grouped by line, and linear regression is applied to each group of points to estimate optimal line parameters for accurate representation of olive tree rows in the image.

After tracing the initial lines, post-processing is performed to improve olive tree row detection. This involves adding lines where wide spacing between rows are identified to ensure a periodic representation of the olive tree row arrangement.

\item \textbf{Detecting olive tree columns}
We use a similar method to trace olive tree rows but focus on identifying possible columns. Our goal is to combine points from different rows to form potential columns. Distances between points from successive rows are calculated, and a column corresponds to the minimal distance between the considered rows. A line is then drawn connecting each pair of points.

To trace all correct columns, the same technique as for rows is used. Histograms are calculated for each possible column, and the one with the maximum variance is identified to determine the correct column. Next, the distances between columns are measured, and all correct columns are traced.

\item \textbf{Locating olive trees through intersection detection}
After the rows and columns of olive trees are traced, the intersections where they cross are found to locate olive trees. In Fig. \ref{fig:intersections} the intersection are shown and almost all non-detected trees are within these intesections.

\end{itemize}

\subsection{Phase 3 : Processing unsegmented olive trees}
Although the SAMG works well for segmenting most olive trees, some may be missed. It is important to identify and process these unsegmented olive trees to ensure all olive trees in the image are accurately segmented.
\begin{itemize}
\item \textbf{Using SAM Predictor with unsegmented olive tree centers}
In this processing, SAM Predictor with points is used, representing the centers of unsegmented olive trees, as input. The predictor analyzes these points and generates masks based on both local and global image features.

The same filtering function with previously established size thresholds (i.e. phase 1) is used to keep only relevant olive tree masks. Points not meeting these criteria are held for further processing. Fig. \ref{fig:SP method 1} shows that many non-detected trees are found using the processing.

\begin{figure}[H]
    \centering
    \includegraphics[width=0.6\linewidth]{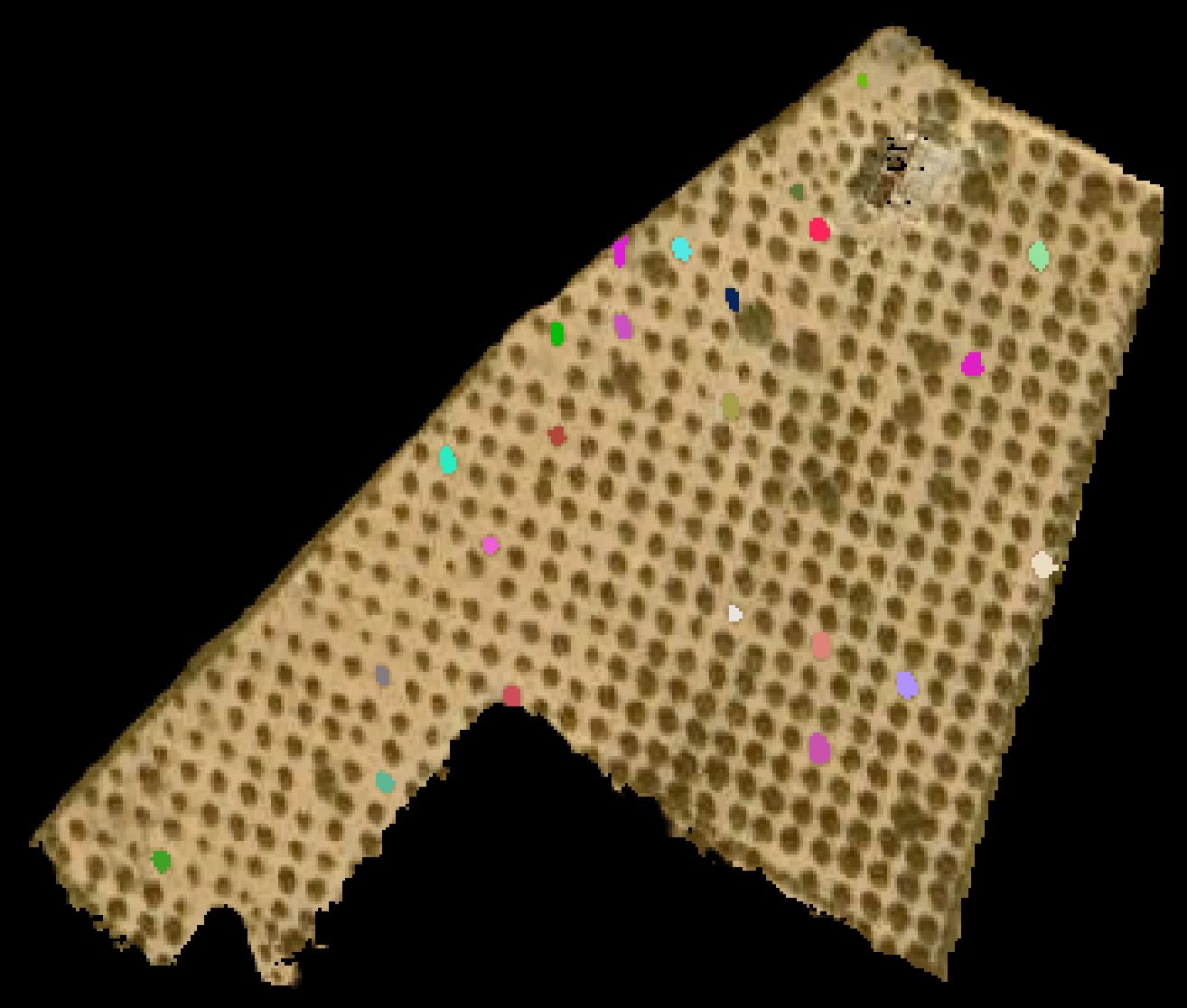}
    \caption{Result of segmenting olive trees using SAM Predictor with centers of unsegmented trees}
    \label{fig:SP method 1}
\end{figure}
\item \textbf{Using SAM Predictor with unsegmented olive tree centers and their bounding boxes}
The remaining points after the second segmentation are those not covered by masks from SAMG or SP. These represent unsegmented olive tree positions that need an additional processing to integrate them into the overall segmentation process.

In this method, bounding boxes are calculated around these points, and they are utilized, along with the points themselves, as input for SP to generate potential masks. Fig. \ref{fig:SP method 2} shows that again many non-detected trees are found using the processing.
\begin{figure}[H]
    \centering
    \includegraphics[width=0.6\linewidth]{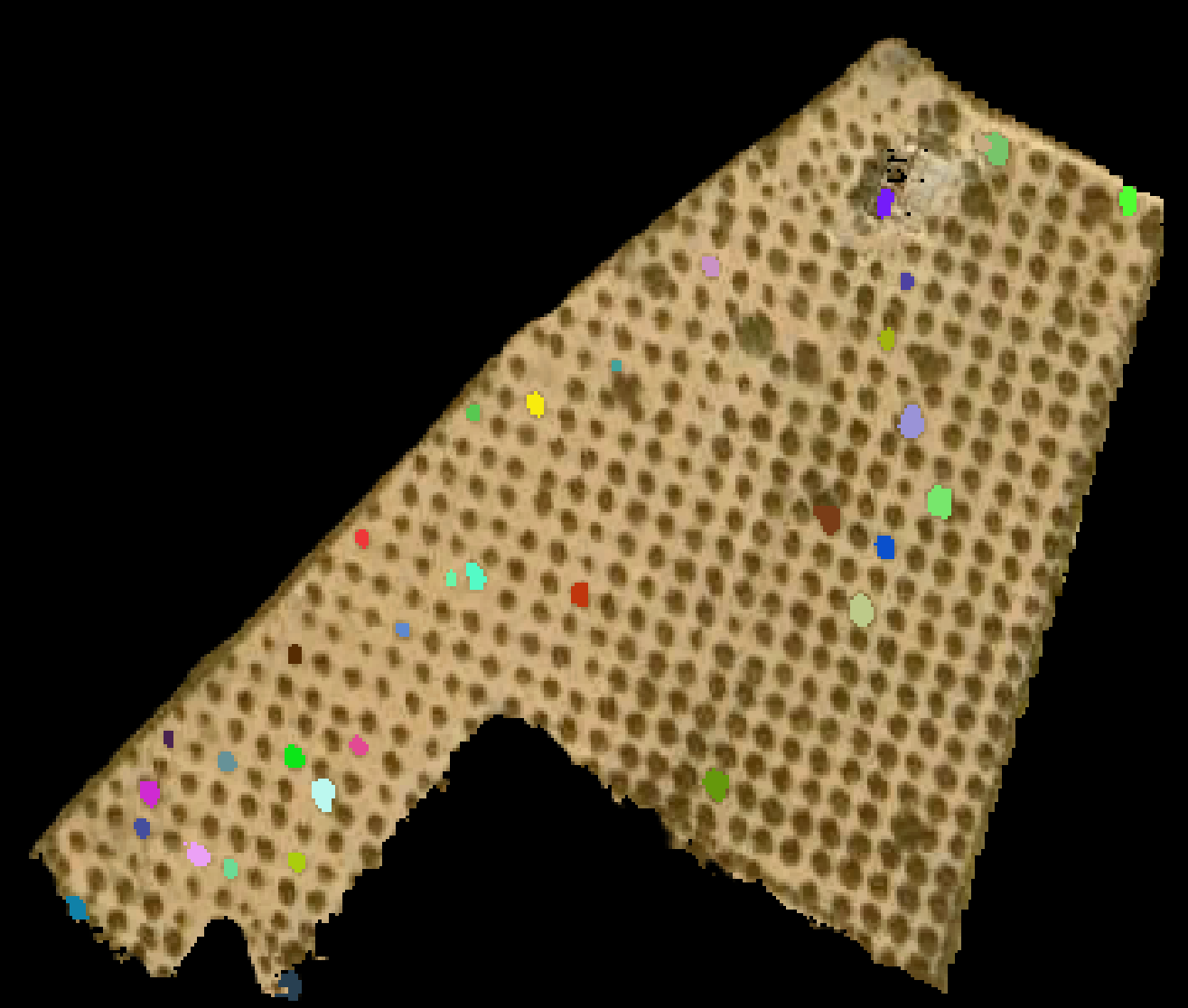}
    \caption{Result of segmentation using SAM Predictor with unsegmented tree centers and their bounding boxes}
    \label{fig:SP method 2}
\end{figure}
\end{itemize}
\subsection{Phase 4: Olive trees filtering}

As segmented trees in the first phase are assumed to be correct, their forms will be used here to filter the trees detected in the third phase of our processing. Indeed, to ensure the correctness in mask selection, we propose to compare the predicted olive tree masks to those generated by SAMG in the initial phase. The masks are filtered using shape filtering, specifically cosine similarity [cf. (\ref{cossim})], to retain only closely matched masks.

\section{Results}
\subsection{Images dataset}
\begin{figure}[H]
    \centering
    \includegraphics[width=0.97\linewidth]{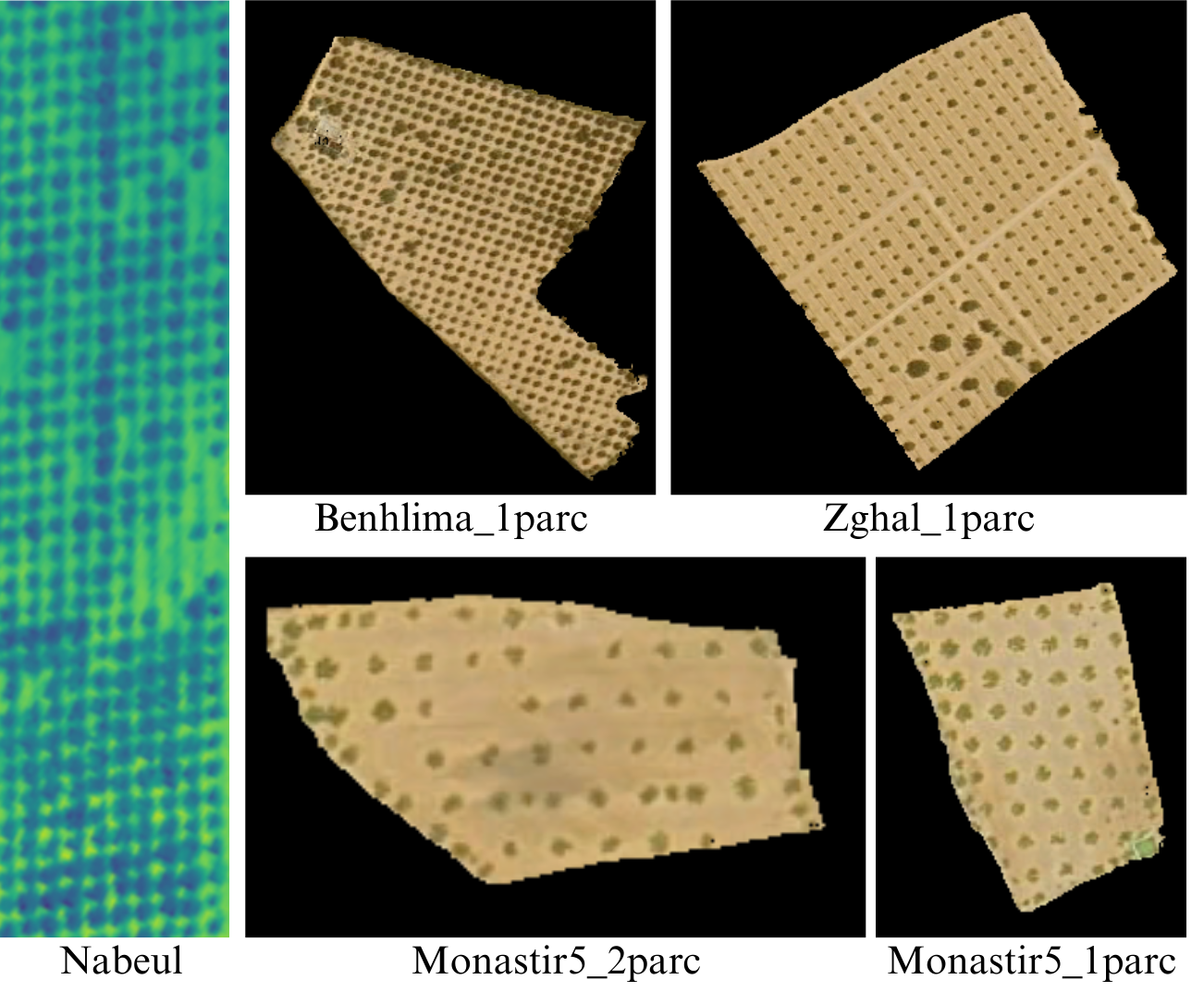}
    \caption{Images dataset}
    \label{fig:used images}
\end{figure}

\begin{figure}[H]
    \centering
    \includegraphics[width=0.97\linewidth]{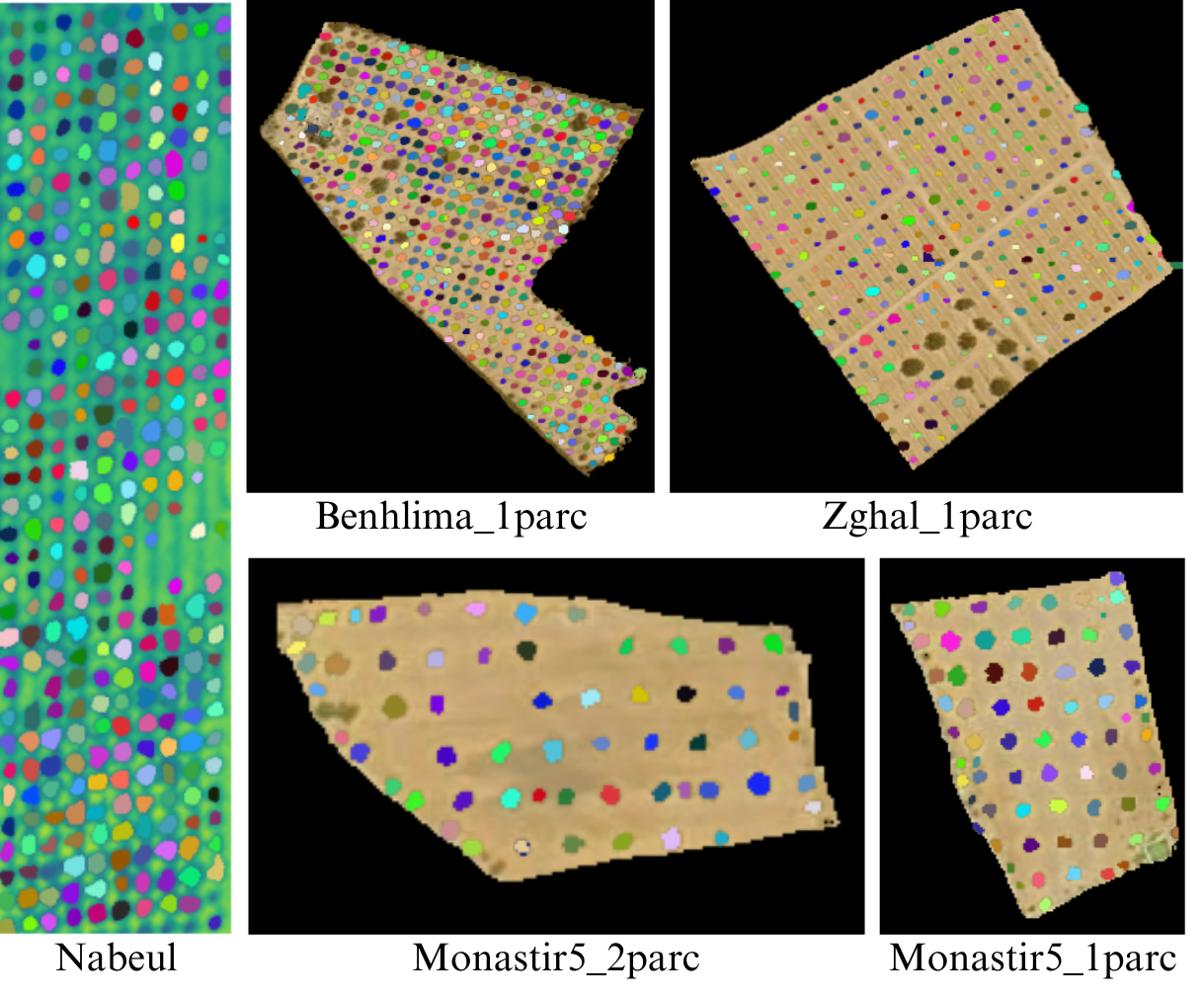}
    \caption{Final results}
    \label{fig:final results}
\end{figure}

For our tests, a set of five images is used, including five satellite images of olive tree plots located in Tunisia [cf. Fig. \ref{fig:used images}].
The segmented images are presented in Fig.~\ref{fig:final results}, which demonstrate visually effective performance. Further details will be discussed in the following section.

\subsection{Performances}

\begin{table*}[t]
\caption{Performance Evaluation}
\begin{center}
\footnotesize
\renewcommand{\arraystretch}{2.8}
\begin{tabular}{|c|c|c|c|c|c|c|}
\hline
\textbf{Plot} & \textbf{True Count} & \textbf{\shortstack{\\ Segmented \\ Olive Tree \\ Count from \\ phase 1}} & \textbf{\shortstack{Count of \\ Predicted Olive \\ Tree Positions \\ phase 2}} & \textbf{\shortstack{Segmented \\ Olive Tree \\ Count from \\ phase 3 - Method 1}}& \textbf{\shortstack{Segmented \\ Olive Tree \\ Count from \\ phase 3 - Method 2}} & \textbf{\shortstack{Total Count \\ of Segmented \\ Olive Trees}}  \\ \hline
Nabeul & 330 & \shortstack{161 \\ 48.78\%} & 349 & 152 & 16   & \shortstack{327 \\ 99.09\%}   \\ \hline
Monastir5\_1parc & 71 & \shortstack{70 \\ 98.59\%} & 71 & 3 & 7 & \shortstack{71 \\ 100\%} \\ \hline
Monastir5\_2parc & 61 & \shortstack{59 \\ 96.72\%} & 82 & 2 & 8 & \shortstack{60 \\ 98.36\%} \\ \hline
Zghal\_1parc & 373 & \shortstack{294 \\ 78.82\%} & 386 & 97 & 1 & \shortstack{369 \\ 98.92\%} \\ \hline
Benhlima\_1parc & 508 & \shortstack{466 \\ 91.73\%} & 523 & 23 & 34 & \shortstack{490 \\ 96.45\%} \\ \hline
Accuracy rate &  & 82.928\% &  &  &  & 98.564\% \\ \hline
\end{tabular}
\end{center}
\label{tab:performance}
\end{table*}

The performance of olive tree segmentation is evaluated by comparing the count of segmented trees with the true count of trees in each plot.\par
\begin{equation}
\text{Performance Evaluation} = \frac{\text{ Count of Segmented Olive Trees}}{\text{True Count of Olive Trees}}
\end{equation}
This ratio allows us to quantify the accuracy of the segmentation. In Table ~\ref{tab:performance}, the results of this evaluation are presented for several plots.

\begin{itemize}
    \item \textbf{Interpretation of phases}
   The segmented olive tree count from phase 1 generally shows results lower than the true counts. There is a notable number of olive trees that are not segmented, particularly for larger plots like ``Nabeul" and ``Zghal", which contain a significant number of olive trees. For that reason, we apply our multi-phase approach to improve the results of SAM. \par
    In phase 2, the approach involves predicting the locations of all olive trees by detecting intersections between rows and columns. The count of these positions is generally higher than the true count, as there are olive trees that are not really present in their designated positions (e.g. uprooted trees). These predicted positions serve as valuable input for the following phase, contributing to enhancing the performance of the first phase. \par
    Phase 3 addresses unsegmented olive trees using two processings. Processing 1 often indicates a higher number of unsegmented olive trees, especially in cases where initial segmentation was less effective. Processing 2 generally results in reduced counts, as it serves as a refinement step aimed at improving the accuracy of segmentation.\par 
    The total count of segmented olive trees for each plot is then calculated, demonstrating improved accuracy after all phases are applied.\par
    
    \item \textbf{Comparative performance analysis }
    Our multi-phase approach achieves an accuracy rate of 98.56\%, surpassing SAM's 82.92\%, as represented by the segmented olive tree count from phase 1. This comparison highlights the limitations of SAM, particularly when dealing with complex scenarios involving low-resolution or detailed images. Conversely, it underscores the effectiveness of our proposed approach in accurately segmenting olive trees from satellite images, especially in challenging plots like ``Nabeul" and ``Zghal", which host many olive trees.
    
\end{itemize}

\section{Conclusion}
In conclusion, our study presented an innovative approach to olive tree segmentation, leveraging advanced techniques including the Segment Anything Model (SAM). Through key steps such as mask generation, detection of olive tree rows and columns, and additional refinements, our method achieved a remarkable 98\% accuracy rate, surpassing the performance of SAM, which achieved 82\%. This advancement holds the potential for precise monitoring of olive tree health, providing invaluable insights for efficient agricultural management and the conservation of essential resources.

Looking ahead, several promising avenues for future work emerge. Exploring fine-tuning of the SAM could further enhance its performance, ultimately leading to even greater accuracy and applicability.
Additionally, our approach could be extended to other types of agriculture that exhibit alignment patterns, such as almond orchards. This would test the versatility and robustness of our method across different crop types.

\bibliographystyle{IEEEtran}
\bibliography{IEEEabrv,references}
\end{document}